\documentclass[runningheads]{llncs}

\usepackage{makecell}
\usepackage[T1]{fontenc}
\usepackage{graphicx}
\usepackage{float}
\usepackage{pgfplots}
\usepackage{pgfplotstable}
\usepackage{subcaption}
\usepackage{amsmath,amssymb}
\usepackage{booktabs}
\usepackage{multirow}
\pgfplotsset{compat=1.18}
\usetikzlibrary{patterns,positioning}

\begin{document}

\title{Noise-Robust Topology Estimation of 2D Image Data via Neural Networks and Persistent Homology}
\titlerunning{Noise-Robust Topology Estimation}

\title{Noise-Robust Topology Estimation of 2D Image Data via Neural Networks and Persistent Homology}
\titlerunning{Noise-Robust Topology Estimation}

\author{Dylan Peek\inst{1}\and
Matthew P. Skerritt\inst{2}\and
Stephan Chalup\inst{1}}

\institute{The University of Newcastle, Newcastle NSW 2308, Australia \and
The Royal Melbourne Institute of Technology, Melbourne VIC 3000, Australia}

\authorrunning{Peek et al.}

\maketitle

\begin{abstract}
Persistent Homology (PH) and Artificial Neural Networks (ANNs) offer contrasting approaches to inferring topological structure from data. In this study, we examine the noise robustness of a supervised neural network trained to predict Betti numbers in 2D binary images. We compare an ANN approach against a PH pipeline based on cubical complexes and the Signed Euclidean Distance Transform (SEDT), which is a widely adopted strategy for noise-robust topological analysis. Using one synthetic and two real-world datasets, we show that ANNs can outperform this PH approach under noise, likely due to their capacity to learn contextual and geometric priors from training data. Though still emerging, the use of ANNs for topology estimation offers a compelling alternative to PH under structural noise.

\end{abstract}

\section{Introduction}
Understanding the topology of data in the presence of noise is a central challenge across many scientific domains, including biology, materials science, and neuroscience. Among the tools developed to infer topological structure, persistent homology (PH)~\cite{OtterEtAl2017} and artificial neural networks (ANNs) represent two fundamentally different paradigms. This study aims to bridge topological data analysis (TDA) and machine learning, and to provide further support for viewing TDA as a meaningful application domain for modern ANN-based methods.

PH is a foundational approach in TDA offering mathematical techniques for extracting topological invariants such as connected components and holes with proven stability under bounded perturbations~\cite{Cohen-SteinerEtAl2007}. While PH is well-suited for real-world applications and includes mechanisms to handle noise in input data, it remains sensitive to outliers, choice of scale parameters, and sampling irregularities~\cite{ChazalMichel2021,FasyEtAl2014}. For overviews on TDA, we refer the reader to~\cite{EdelsbrunnerHarer2010,OtterEtAl2017,HenselEtAl2021}.

In contrast, previous studies show that ANNs such as convolutional neural networks (CNNs) offer data-driven models that can implicitly learn topological features from raw input such as images or voxel data~\cite{HannouchChalup2025,paul2019estimating,PeekEtAl2023}.

This result motivates a broader hypothesis: ANNs trained for topology estimation may, in some cases, be more robust to noise than a traditional PH approach. ANNs can leverage contextual and geometric priors such as texture, curvature, or spatial patterns, that can co-occur with topological features. For example, if holes in a material tend to have a characteristic size, shape, or surrounding texture, ANNs can learn these associations. While PH can reflect some metric information of the data in persistence diagrams, it usually lacks direct access and analytical capacity of the full geometric context in the input dataset.

In our experiments, we compare the robustness of PH and an ANN-based approach for estimating the topology of 2D data under varying noise conditions. Using three datasets, including two from real-world sources, we observe that ANNs can outperform PH with Signed Euclidean Distance Transform (SEDT) in noisy settings. To support this comparison, we include a newly generated synthetic dataset that enables controlled testing of topological robustness. The analysis highlights the differences in how each method responds to increasing noise and reveals practical limitations of the PH approach under real-world conditions. The results also inform the design of noise-resilient ANN architectures for topology estimation. The discussion aims to foster dialogue between the largely separate communities working in TDA and machine learning.

\section{Background and Related Work}

PH is a topological data analysis technique that tracks the appearance and disappearance of features such as path-connected components and loops across a range of filtration parameters. In 2D images, Betti numbers provide interpretable descriptors: $\beta_0$ counts connected components, and $\beta_1$ counts topological loops or holes (cf. Fig.~\ref{fig:ph_barcode_example}). By constructing a filtration (an increasing sequence of nested topological spaces) PH quantifies the persistence of each feature, enabling robust, multi-scale analysis that can distinguish signal from noise~\cite{singh2023topological}.

The effectiveness of PH depends heavily on the filtration strategy and the complex used. While grayscale images often permit intensity sublevel filtrations~\cite{chung2024morphological}, binary images require geometry-aware approaches. One such approach is the SEDT, which assigns each binary pixel a signed distance to the nearest opposite pixel. A sublevel filtration on the SEDT function progressively fills each hole from its center outward before including the surrounding structure, producing persistence diagrams where prominent $\beta_1$ features correspond to geometric holes~\cite{pritchard2023persistent}. This method not only preserves hole size information but also generalizes to 3D.

Compared to other strategies such as morphological filtrations~\cite{chung2024morphological} or distance-to-measure (DTM) filtrations~\cite{turkevs2021noise}, SEDT offers a principled, shape-centric framework for analyzing binary masks with topological noise tolerance. However, PH on real data still faces practical challenges: spurious features due to segmentation artifacts, inconsistent scale between samples, and the difficulty of extracting Betti numbers from continuous persistence diagrams.

In applications and case studies, 
PH-based hole counting has been applied across biomedical and materials imaging. In biomedicine, Pritchard \emph{et al.}~\cite{pritchard2023persistent} used PH with an SEDT filtration to quantify microscopic pores in bone tissue, revealing significantly higher cortical porosity in mutant mice.
PH has also been used in tumor histology to analyze cell morphology and immune infiltration. %
Others linked loop-like spatial features to patient survival using persistence landscapes and distance-based filtrations~\cite{moon2020predicting,vipond2021multiparameter}.
Additional medical studies include skin barrier assessment using Manhattan-distance-PH~\cite{koseki2020assessment}, and various applications to gland segmentation and vasculature analysis.
In materials science, PH enables pore network characterization in 2D and 3D. Robins \emph{et al.}~\cite{robins2016percolating} related persistent Betti numbers in micro-CT images of rock to percolation thresholds, while Herring \emph{et al.}~\cite{herring2019topological} identified trapped fluid clusters in sandstone.%

These examples underscore PH’s versatility in quantifying holes across domains. However, image-based PH pipelines require careful tuning of filtration strategies and thresholds, and interpreting persistence diagrams remains application-specific and sensitive to scale, noise, and segmentation artifacts.

The use of ANNs to estimate Betti numbers from simulated data was first demonstrated by Paul \emph{et al.}~\cite{paul2019estimating}, who employed a basic CNN architecture on examples of 2D and 3D point cloud data. Subsequent work explored alternative ANN architectures, focusing on 3D data~\cite{PeekEtAl2023}. The case of topological 4D image-type data, including downscaling issues, was addressed in~\cite{HannouchChalup2023,HannouchChalup2025}. These studies emphasize performance and efficiency benefits of ANN-based estimation. The use of ANNs for TDA motivated our empirical study on noise-specific training and evaluation over progressively distorted images.

\subsection{Noise Robustness in Topological Estimation}

Recent work has begun addressing the limitations of PH in noisy or perturbed settings. RipsNet~\cite{de2022ripsnet} proposes a neural architecture that approximates persistence diagrams from point cloud data while providing theoretical robustness to localized perturbations. By bounding the Wasserstein distance between predicted and true diagrams under fractional point corruption, RipsNet mitigates outlier sensitivity inherent in Rips complexes. However, this robustness is inherently local, designed for sparse per-point noise, and does not address broader structural distortions or dense input degradation, especially in image-based data.

Heiss \emph{et al.}~\cite{heiss2021impact} study the effect of resolution changes on PH in 2D and 3D grayscale images. They demonstrate that aggressive downsampling can erase small topological features, motivating tradeoffs between fidelity and computational feasibility. As image resolution increases, both the SEDT and cubical filtration grow in cost, often superlinearly. We note that, in contrast, CNNs scale linearly with input size (given fixed kernel and stride), offering a computational advantage at high resolution.

Our study builds on these insights by training CNNs to estimate topological descriptors from images under severe structural corruption. Unlike RipsNet, which estimates the topology of perturbed inputs, our model infers the \emph{original} topology from broken inputs, learning to reconstruct meaningful topological summaries from noise. This includes degradation that affects local connectivity, hole closure, or texture, which traditional PH pipelines struggle to interpret. Moreover, our approach retains scalability across resolutions without sacrificing fidelity, offering a data-driven alternative to explicit filtration pipelines in high-noise regimes.

\section{Datasets}

This study utilises three datasets: a synthetic Voronoi-based set created as part of this study, a semi-synthetic materials dataset DeePore~\cite{rabbani2020deepore}, and the real-world cellular electron microscopy dataset CEM500K~\cite{conrad2021cem500k}. Each dataset is processed into clean, binary 2D images where foreground regions represent solid material and background corresponds to void space. Prominent topological structures such as connected components ($\beta_0$) and holes ($\beta_1$) are then extracted.

We generated the synthetic Voronoi dataset to control for holes and structure quantities in an unambiguous, noise-free state. Voronoi images were generated by randomly sampling 2D points and clustering them with $k$-means~\cite{Lloyd1982}. Each cluster undergoes Voronoi tessellation~\cite{Aurenhammer1991} to form region masks. A global Perlin noise mask~\cite{Perlin1985} modulates edge thickness across regions to introduce spatial variability. 
These samples emulate material microstructure, exhibiting a controlled yet diverse range of topologies. Ground-truth Betti numbers are calculated before any noise is applied.

DeePore~\cite{rabbani2020deepore} is a volumetric dataset of simulated porous media, designed to capture realistic transport and morphology patterns. We slice each $256^3$ volume axially into three 2D grayscale images (top, middle, bottom), producing 53,100 total samples.

CEM500K~\cite{conrad2021cem500k} contains grayscale electron microscopy images of subcellular environments. From an initial pool of 100,000 slices, we retain a subset of 61,177 that exhibit topological diversity and share a common resolution.

In both DeePore and CEM500K datasets, each image is binarized using Otsu thresholding~\cite{Otsu1979}, then cleaned with morphological opening and closing (disk size 3)~\cite{Serra1982} to remove small specks and seal fine voids. The result is inverted to treat pores as foreground before labeling with cubical PH (see overview in Table~\ref{tab:dataset_summary}).

\label{sec:datasets}
\begin{table}[h]
\centering
\caption{Summary of datasets used in this work. Each clean image is augmented with five noise variants, yielding a total of 698{,}885 images.}
\begin{tabular}{lccccc}
\toprule
\textbf{Dataset} & $\boldsymbol{\beta_0}$ \textbf{Range} & $\boldsymbol{\beta_1}$ \textbf{Range} & \textbf{Clean Samples} & \textbf{Image Size} & \textbf{Total Images} \\
\midrule
Voronoi   & 1--5   & 0--50  & 25,500  & $512 \times 512$ & 127,500 \\
DeePore  & 1--98  & 0--24  & 53,100  & $256 \times 256$ & 265,500 \\
CEM500K   & 1--73  & 0--50  & 61,177  & $224 \times 224$ & 305,885 \\
\midrule
\textbf{Total}    & --     & --     & \textbf{139,777} & --         & \textbf{698,885} \\
\bottomrule
\end{tabular}
\label{tab:dataset_summary}
\end{table}

\begin{figure}[h]
\centering
\setlength{\tabcolsep}{2pt}
\renewcommand{\arraystretch}{1}
\begin{tabular}{lccc}
& \textbf{Voronoi} & \textbf{DeePore} & \textbf{CEM500K} \\\\

N0 &
\includegraphics[width=0.27\linewidth, height=0.27\linewidth]{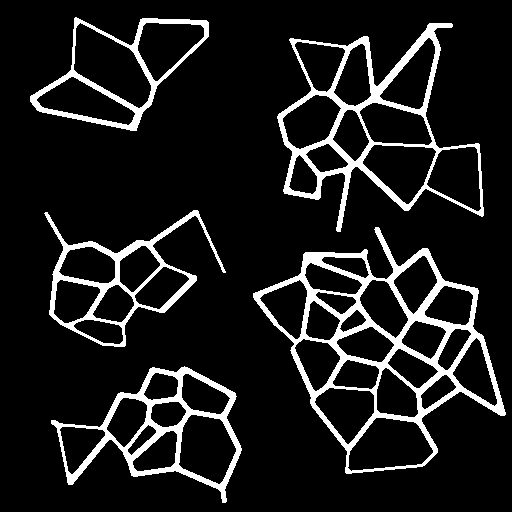} &
\includegraphics[width=0.27\linewidth, height=0.27\linewidth]{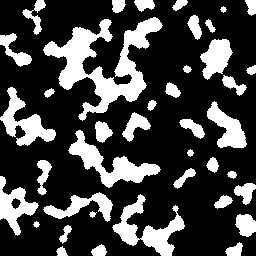} &
\includegraphics[width=0.27\linewidth, height=0.27\linewidth]{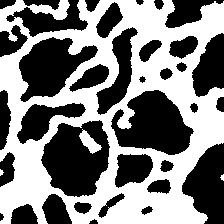} \\

N2 &
\includegraphics[width=0.27\linewidth, height=0.27\linewidth]{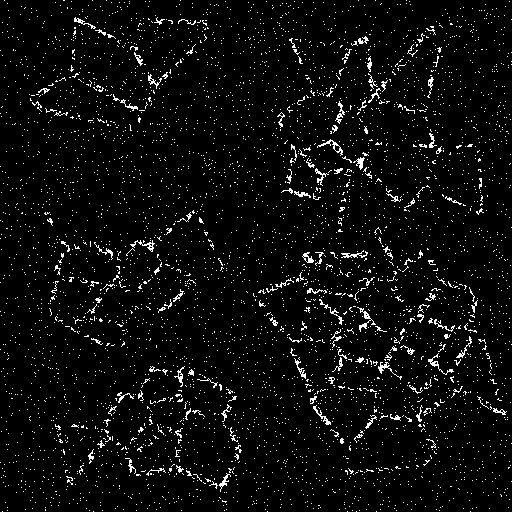} &
\includegraphics[width=0.27\linewidth, height=0.27\linewidth]{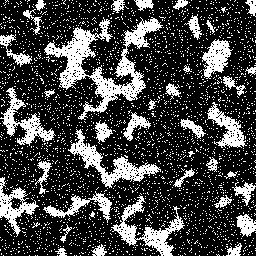} &
\includegraphics[width=0.27\linewidth, height=0.27\linewidth]{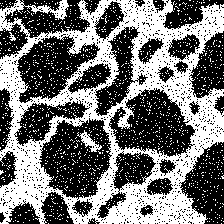} \\

N4 &
\includegraphics[width=0.27\linewidth, height=0.27\linewidth]{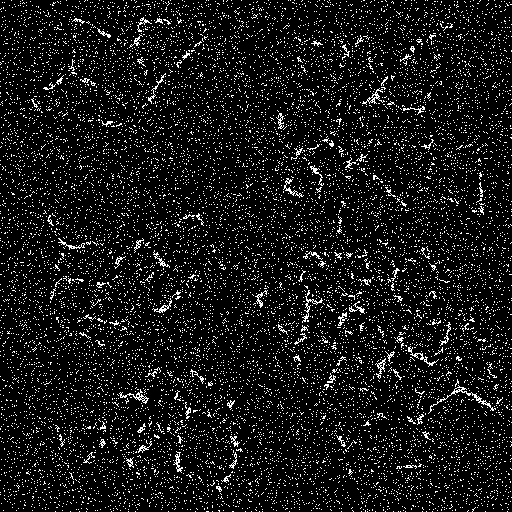} &
\includegraphics[width=0.27\linewidth, height=0.27\linewidth]{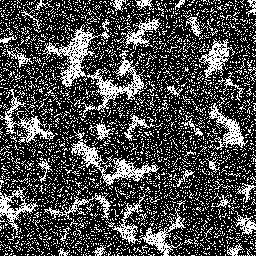} &
\includegraphics[width=0.27\linewidth, height=0.27\linewidth]{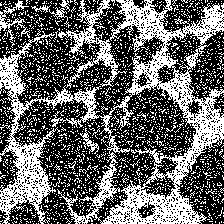} \\\\

\end{tabular}
\caption{Samples from each dataset (columns) shown across 3 noise levels (rows, N0, N2, N4), with N0 representing clean images and N4 representing highest distortion. The datasets are used under the CC BY 4.0 license (https://creativecommons.org/licenses/by/4.0/).}
\label{fig:dataset_noise_matrix}
\end{figure}

\begin{table}[h]
\centering
\caption{Noise configuration used at each distortion level (N1–N4) for each dataset. 
Both additive and subtractive forms of Gaussian and Perlin noise were applied, simulating spatially localized corruption and structural degradation. 
Note that while parameters are fixed, the perceptual impact varies by dataset resolution: higher-resolution images are more resilient to a given noise level than smaller images.}
\label{tab:noise_summary}
\begin{tabular}{@{}lcccccc@{}}
\toprule
\textbf{Dataset} & \textbf{Level} & \makecell[c]{Edge\\Peeling} & \makecell[c]{Gaussian Noise\\(mean, $\sigma$)} & \makecell[c]{Perlin\\Scale} & \makecell[c]{Perlin\\Threshold} \\
\midrule
\multirow{4}{*}{Voronoi} 
& N1 & -- & (0, 5.0)  & 0.125 & -- \\
& N2 & -- & (0, 10.0) & 0.10  & -- \\
& N3 & -- & (0, 15.0) & 0.075 & -- \\
& N4 & -- & (0, 20.0) & 0.05  & -- \\
\midrule
\multirow{4}{*}{DeePore} 
& N1 & 1× (50\%) & (–3.0, 2.0) & 0.10 & 0.40 \\
& N2 & 2× (50\%) & (–2.5, 2.0) & 0.50 & 0.35 \\
& N3 & 3× (50\%) & (–2.0, 2.0) & 0.50 & 0.25 \\
& N4 & 4× (60\%) & (–1.5, 2.0) & 0.50 & 0.175 \\
\midrule
\multirow{4}{*}{CEM500K} 
& N1 & 1× (50\%) & (–3.5, 2.0) & 0.10 & 0.40 \\
& N2 & 2× (50\%) & (–3.0, 2.0) & 0.50 & 0.35 \\
& N3 & 3× (50\%) & (–2.5, 2.0) & 0.50 & 0.25 \\
& N4 & 4× (60\%) & (–2.0, 2.0) & 0.50 & 0.175 \\
\bottomrule
\end{tabular}
\end{table}

To evaluate robustness under degradation, we apply synthetic noise at five discrete levels (N0–N4) after binarization. Importantly, noise is introduced after labeling the clean, preprocessed images to preserve the N0 topology as ground truth. This ensures that performance metrics reflect only the effect of added noise. The morphological cleaning step is not part of the proposed TDA pipeline, but serves as an experimental control to isolate the impact of noise on topological estimation.

Each noisy variant is generated by applying a combination of Perlin and Gaussian noise, and random boundary erosion. See Table~\ref{tab:noise_summary} for noise parameters.

\section{Methodology}
Each dataset was partitioned into 70\% training, 15\% validation, and 15\% test sets for ANN training and testing. PH, which alone is not a machine learning method, was evaluated on the same test sets for comparison. 
When referring to the dataset names within methodology and results, we are referring to the preprocessed, noise induced samples outlined in Section~\ref{sec:datasets}.

\subsection{Persistent Homology Baseline}\label{subsec:ph_method}
\begin{figure}[ht]
\centering
\begin{minipage}{0.35\linewidth}
    \centering
    \includegraphics[width=\linewidth]{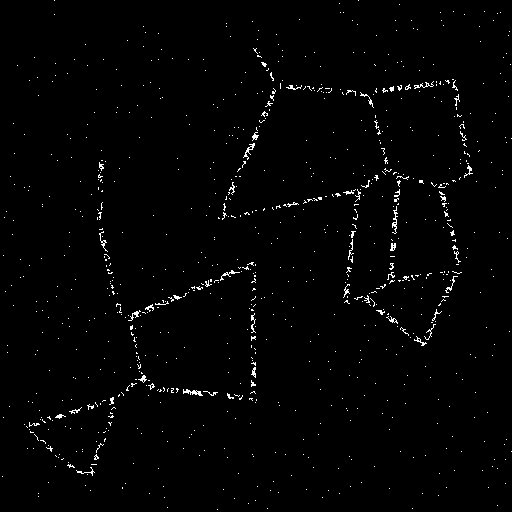}
\end{minipage}
\hfill
\begin{minipage}{0.55\linewidth}
    \centering
    \includegraphics[width=\linewidth]{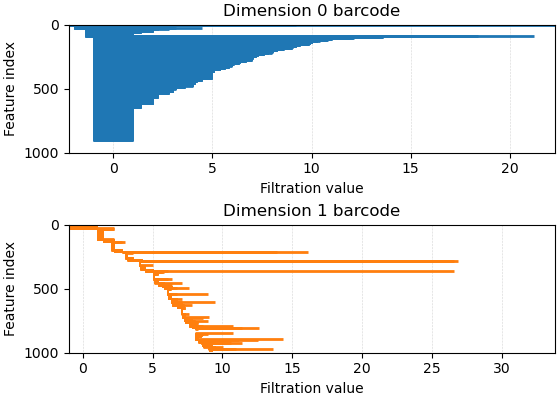}
\end{minipage}
\caption{Example image and its corresponding persistence barcode computed using SEDT filtration. The lines indicate the birth and death of features as the sub-level filtration increases. Long bars indicate the existence of features ($\beta_0$ = number of path-connected components; $\beta_1$ = number of loops) across a larger scale, while short bars may represent noise. In a noise-free situation this example would result in Betti numbers $\beta_0$ = 2 and $\beta_1$ = 7.}
\label{fig:ph_barcode_example}
\end{figure}

We construct a PH pipeline using the SEDT. The resulting scalar fields are analyzed via sublevel-set filtrations using cubical complexes, enabling the computation of persistence diagrams for homology dimensions $H_0$ and $H_1$—representing connected components and loops, respectively. 
The persistence barcodes in Figure~\ref{fig:ph_barcode_example} show the birth and death times of features as they are formed throughout this filtration~\cite{pritchard2023persistent}.
To extract discrete Betti numbers from the persistence information, we employ a windowing strategy based on three parameters: (1) a birth-time lower bound, (2) a birth-time upper bound, and (3) a minimum persistence threshold. A feature is counted only if its birth falls within the defined birth window and its lifespan exceeds the specified threshold~\cite{barnes2021comparative}. This strategy encapsulates other standard persistence thresholding methods (such as purely thresholding lifespans) and allows flexible isolation of significant features based on their location in the birth–death plane.

To provide an upper-bound comparison for our learned models, we conduct an independent grid search for each dataset, noise level, and homology dimension. This creates 90 unique configurations (3 datasets $\times$ 5 noise levels $\times$ 2 Betti dimensions), each optimized separately. For every configuration, a dense $25^3$ grid search (15{,}625 total combinations) is performed over the three windowing parameters. This scale aligns with prior work that emphasizes the importance of fine-grained parameter tuning in topological pipelines~\cite{turkevs2021noise}.

Performance is evaluated using the mean absolute error (MAE) between predicted and pre-noise ground-truth Betti numbers, quantifying the average prediction deviation. This setup, while exhaustive, ensures a maximally calibrated PH pipeline per dataset and noise condition. In contrast, our neural network model is trained and evaluated jointly across all noise levels, highlighting the advantage of contextual generalization and noise-adaptive inference.

It is worth noting that our PH implementation represents one well-established configuration to provide a reference point for ANN performance. Many alternatives exist in terms of filtration strategies (e.g., distance-to-measure, morphological), complex types (e.g., Vietoris–Rips, alpha complexes), and preprocessing steps (e.g., denoising, thresholding). While these decisions can be tuned to improve robustness in specific applications, such tailoring requires expert domain and TDA knowledge. In contrast, our neural models require no dataset-specific configuration, learning relevant topological features and their noise characteristics directly from training data.
\subsection{Neural Network}
\label{subsec:nn_method}

We adopt a supervised learning approach to estimate Betti numbers directly from binary images. After initial benchmarking with conventional convolutional architectures, including ResNet and basic CNN variants, the ConvNeXt-Large architecture~\cite{liu2022convnet} was selected as the backbone due to its superior performance.

The output head was modified for regression, producing two scalar values corresponding to the predicted Betti numbers $\beta_0$ and $\beta_1$. A standard mean squared error (MSE) loss was used during training. To stabilize training across datasets with differing topological scales, output targets were normalized to the range $[0, 1]$ based on dataset label statistics: $\beta_0 \in [1, 50]$ and $\beta_1 \in [0, 100]$. Predictions were denormalized before evaluation.

To emulate real-world deployment scenarios involving staged domain adaptation, we trained the model in three successive stages. In the first stage, the model $\text{NN}_\text{V}$ was trained on the synthetic Voronoi dataset. In the second stage, it was fine-tuned on the DeePore dataset, resulting in $\text{NN}_\text{VD}$, and finally adapted to CEM500K, yielding $\text{NN}_\text{VDC}$. This progression reflects a transition from simulated to increasingly complex and heterogeneous real-world geometries. An additional model $\text{NN}_\text{Simul}$ was trained once on all datasets simultaneously. Throughout the remainder of the paper, we use this subscript notation to denote each model variant, with subscripts indicating the datasets used for training. The network was trained on all noise levels simultaneously, unlike PH, which was optimally calibrated per noise level. 
During each training epoch, the samples experienced data augmentation to increase generalization and minimize overfitting. This included random rotation and mirroring. 

Training used the Adam optimizer with an initial learning rate of $1 \times 10^{-3}$ and cosine annealing scheduler. The fine-tuning phases on downstream datasets used reduced learning rates of $1 \times 10^{-4}$ and $1 \times 10^{-5}$. Each model was trained with early stopping based on validation loss, and the final test performance is reported in terms of MAE and standard deviation, using a fixed test set. Ground-truth labels for all noise variants were taken from the corresponding clean image.

\section{Results}
\begin{table}[h!]
\centering
\caption{MAE and STD for predicted Betti numbers $\beta_0$ and $\beta_1$ across noise levels N0–N4. NN model subscripts indicate training sets, and multiple subscript characters indicate transfer learning sequences across datasets. The bold labelled rows on the left highlight the most relevant results and are visualised in Figure~\ref{fig:mae_betti_plots}.}
\label{tab:mae_std_results}
\scriptsize
\setlength{\tabcolsep}{2.5pt}
\renewcommand{\arraystretch}{1.1}

\begin{tabular}{llccccc}
\toprule
& \textbf{Model} & \textbf{N0} & \textbf{N1} & \textbf{N2} & \textbf{N3} & \textbf{N4} \\
\midrule

\multicolumn{7}{c}{\textbf{Voronoi – Betti $\beta_0$}} \\
           & \textbf{PH}         & 0.006 ± 0.076 & 0.154 ± 0.378 & 0.766 ± 0.833 & 1.416 ± 1.176 & 1.617 ± 1.213\\
           & \textbf{$\text{NN}_\text{V}$}   & 0.130 ± 0.105 & 0.165 ± 0.157 & 0.176 ± 0.174 & 0.215 ± 0.208 & 0.275 ± 0.254 \\
           & $\text{NN}_\text{VD}$  & 1.464 ± 0.749 & 42.998 ± 19.421 & 42.496 ± 19.067 & 37.985 ± 17.444 & 28.189 ± 13.823 \\
           & $\text{NN}_\text{VDC}$ & 1.382 ± 0.986 & 19.246 ± 8.663 & 19.773 ± 8.849 & 17.191 ± 7.994 & 11.692 ± 6.008 \\
           & $\text{NN}_\text{Simul}$ & 0.191 ± 0.168 & 0.196 ± 0.194 & 0.236 ± 0.219 & 0.279 ± 0.250 & 0.342 ± 0.301 \\

\multicolumn{7}{c}{\textbf{Voronoi – Betti $\beta_1$}} \\
           & \textbf{PH}         & 0.000 ± 0.000 & 1.141 ± 1.161 & 3.948 ± 3.376 & 11.309 ± 8.545 & 11.919 ± 8.838\\
           & \textbf{$\text{NN}_\text{V}$}   & 0.096 ± 0.112 & 0.210 ± 0.300 & 0.297 ± 0.377 & 0.498 ± 0.553 & 0.900 ± 0.882 \\
           & $\text{NN}_\text{VD}$  & 18.690 ± 10.873 & 24.840 ± 14.637 & 24.867 ± 14.648 & 24.891 ± 14.650 & 24.934 ± 14.645 \\
           & $\text{NN}_\text{VDC}$ & 19.975 ± 11.763 & 24.884 ± 14.692 & 24.894 ± 14.697 & 24.898 ± 14.704 & 24.904 ± 14.709 \\
           & $\text{NN}_\text{Simul}$ & 0.545 ± 0.407 & 0.365 ± 0.332 & 0.449 ± 0.442 & 0.640 ± 0.618 & 1.054 ± 1.000 \\

\midrule

\multicolumn{7}{c}{\textbf{DeePore – Betti $\beta_0$}} \\
           & \textbf{PH}         & 1.386 ± 1.224 & 1.956 ± 1.799 & 1.628 ± 1.462 & 2.422 ± 1.994 & 11.898 ± 7.813 \\
           & $\text{NN}_\text{V}$   & 26.736 ± 17.306 & 40.142 ± 26.456 & 45.362 ± 28.303 & 58.141 ± 32.176 & 96.664 ± 24.966 \\
           & \textbf{$\text{NN}_\text{VD}$}  & 1.246 ± 1.119 & 1.455 ± 1.429 & 2.490 ± 2.542 & 3.933 ± 4.328 & 5.912 ± 6.455 \\
           & $\text{NN}_\text{VDC}$ & 8.466 ± 4.479 & 9.107 ± 5.172 & 10.679 ± 6.799 & 12.516 ± 8.760 & 12.602 ± 10.427 \\
           & $\text{NN}_\text{Simul}$ & 1.186 ± 0.985 & 1.343 ± 1.340 & 2.371 ± 2.887 & 3.646 ± 4.598 & 5.174 ± 5.835 \\

\multicolumn{7}{c}{\textbf{DeePore – Betti $\beta_1$}} \\
           & \textbf{PH}         & 0.000 ± 0.000 & 0.394 ± 0.723 & 0.412 ± 0.703 & 0.728 ± 0.920 & 1.014 ± 1.524 \\
           & $\text{NN}_\text{V}$   & 17.875 ± 15.332 & 69.627 ± 31.449 & 88.273 ± 36.031 & 119.125 ± 36.204 & 161.171 ± 14.941 \\
           & \textbf{$\text{NN}_\text{VD}$}  & 0.271 ± 0.391 & 0.271 ± 0.429 & 0.371 ± 0.695 & 0.547 ± 0.973 & 0.955 ± 1.677 \\
           & $\text{NN}_\text{VDC}$ & 0.541 ± 0.767 & 0.594 ± 0.846 & 0.699 ± 1.055 & 0.826 ± 1.277 & 1.210 ± 2.060 \\
           & $\text{NN}_\text{Simul}$ & 0.281 ± 0.408 & 0.291 ± 0.617 & 0.401 ± 1.101 & 0.604 ± 1.539 & 0.929 ± 1.592 \\

\midrule

\multicolumn{7}{c}{\textbf{CEM500K – Betti $\beta_0$}} \\
           & \textbf{PH}         & 0.990 ± 1.504 & 2.016 ± 2.266 & 2.624 ± 3.121 & 6.858 ± 6.990 & 8.606 ± 10.029 \\
           & $\text{NN}_\text{V}$   & 7.883 ± 8.046 & 10.887 ± 12.458 & 11.764 ± 13.704 & 60.444 ± 22.025 & 60.215 ± 18.852 \\
           & $\text{NN}_\text{VD}$  & 3.803 ± 3.461 & 4.404 ± 4.059 & 4.664 ± 4.637 & 5.253 ± 5.013 & 5.199 ± 5.102 \\
           & \textbf{$\text{NN}_\text{VDC}$} & 0.585 ± 0.524 & 0.706 ± 0.714 & 0.820 ± 0.923 & 1.360 ± 1.579 & 2.624 ± 3.335 \\
           & $\text{NN}_\text{Simul}$ & 0.499 ± 0.525 & 0.605 ± 0.571 & 0.803 ± 0.759 & 1.448 ± 1.703 & 2.781 ± 3.443 \\

\multicolumn{7}{c}{\textbf{CEM500K – Betti $\beta_1$}} \\
           & \textbf{PH}         & 0.028 ± 0.208 & 0.730 ± 1.918 & 0.572 ± 1.691 & 1.770 ± 4.420 & 1.754 ± 4.423 \\
           & $\text{NN}_\text{V}$   & 11.765 ± 17.102 & 28.867 ± 31.817 & 33.075 ± 36.959 & 158.796 ± 44.028 & 144.214 ± 25.066 \\
           & $\text{NN}_\text{VD}$  & 0.650 ± 1.249 & 0.565 ± 1.080 & 0.486 ± 0.906 & 0.535 ± 0.913 & 0.522 ± 0.874 \\
           & \textbf{$\text{NN}_\text{VDC}$} & 0.161 ± 0.366 & 0.159 ± 0.317 & 0.181 ± 0.371 & 0.221 ± 0.462 & 0.453 ± 1.115 \\
           & $\text{NN}_\text{Simul}$ & 0.225 ± 0.445 & 0.205 ± 0.322 & 0.226 ± 0.391 & 0.317 ± 0.795 & 0.454 ± 0.967 \\

\bottomrule
\end{tabular}
\end{table}

\begin{figure}[h]
\centering
\resizebox{0.475\linewidth}{!}{%
\begin{tikzpicture}
\begin{axis}[
    width=\linewidth,
    height=6cm,
    title={Mean Absolute Error for $\beta_0$},
    xlabel={Noise Level},
    ylabel={MAE},
    xtick={0,1,2,3,4},
    legend style={at={(0.02,0.98)}, anchor=north west, font=\small, draw=none, fill=none},
    grid=both,
    ymin=0, ymax=6,
    ultra thick
]
\addplot[dotted, red, ultra thick]    coordinates {(0,0.006)(1,0.154)(2,0.766)(3,1.416)(4,1.617)};
\addlegendentry{Voronoi PH}
\addplot[solid, red, ultra thick]     coordinates {(0,0.130)(1,0.165)(2,0.176)(3,0.215)(4,0.275)};
\addlegendentry{Voronoi $\text{NN}_\text{V}$}

\addplot[dotted, green!60!black, ultra thick]  coordinates {(0,1.386)(1,1.956)(2,1.628)(3,2.422)(4,11.898)};
\addlegendentry{DeePore PH}
\addplot[solid, green!60!black, ultra thick]   coordinates {(0,1.246)(1,1.455)(2,2.490)(3,3.933)(4,5.912)};
\addlegendentry{DeePore $\text{NN}_\text{VD}$}

\addplot[dotted, blue, ultra thick]   coordinates {(0,0.990)(1,2.016)(2,2.624)(3,6.858)(4,8.606)};
\addlegendentry{CEM PH}
\addplot[solid, blue, ultra thick]    coordinates {(0,0.585)(1,0.706)(2,0.820)(3,1.360)(4,2.624)};
\addlegendentry{CEM $\text{NN}_\text{VDC}$}
\end{axis}
\end{tikzpicture}%
}
\hspace{1em}
\resizebox{0.475\linewidth}{!}{%
\begin{tikzpicture}
\begin{axis}[
    width=\linewidth,
    height=6cm,
    title={Mean Absolute Error for $\beta_1$},
    xlabel={Noise Level},
    ylabel={MAE},
    xtick={0,1,2,3,4},
    legend style={at={(0.02,0.98)}, anchor=north west, font=\small, draw=none, fill=none},
    grid=both,
    ymin=0, ymax=6,
    ultra thick
]
\addplot[dotted, red, ultra thick]    coordinates {(0,0.000)(1,1.141)(2,3.948)(3,11.309)(4,11.919)};
\addlegendentry{Voronoi PH}
\addplot[solid, red, ultra thick]     coordinates {(0,0.096)(1,0.210)(2,0.297)(3,0.498)(4,0.900)};
\addlegendentry{Voronoi $\text{NN}_\text{V}$}

\addplot[dotted, green!60!black, ultra thick]  coordinates {(0,0.000)(1,0.394)(2,0.412)(3,0.728)(4,1.014)};
\addlegendentry{DeePore PH}
\addplot[solid, green!60!black, ultra thick]   coordinates {(0,0.271)(1,0.271)(2,0.371)(3,0.547)(4,0.955)};
\addlegendentry{DeePore $\text{NN}_\text{VD}$}

\addplot[dotted, blue, ultra thick]   coordinates {(0,0.028)(1,0.730)(2,0.572)(3,1.770)(4,1.754)};
\addlegendentry{CEM PH}
\addplot[solid, blue, ultra thick]    coordinates {(0,0.161)(1,0.159)(2,0.181)(3,0.221)(4,0.453)};
\addlegendentry{CEM $\text{NN}_\text{VDC}$}
\end{axis}
\end{tikzpicture}%
}

\caption{Mean Absolute Error (MAE) of Betti number predictions for $\beta_0$ (left) and $\beta_1$ (right) across increasing noise levels.}
\label{fig:mae_betti_plots}
\end{figure}
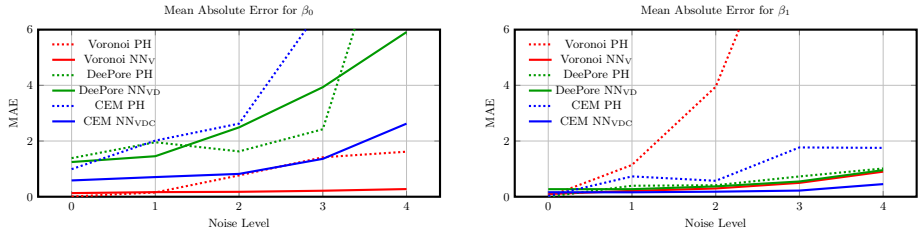

The results in Table~\ref{tab:mae_std_results} summarize the MAE and standard deviation (STD) of predicted Betti numbers $\beta_0$ and $\beta_1$ across five noise levels (N0–N4) for each dataset. MAE is reported in units of absolute count error, reflecting the average difference between predicted and ground-truth Betti numbers. 

Figure~\ref{fig:mae_betti_plots} provides a visual comparison of MAE across noise levels for $\beta_0$ and $\beta_1$ predictions, highlighting both the PH baseline (dotted lines) and the best-performing neural network model for each dataset (solid lines). These plots focus on the models most recently fine-tuned for each dataset. Full evaluations of training stage combinations can be see in Table~\ref{tab:mae_std_results}.

\section{Discussion and Conclusion}
Across all datasets and noise levels, the neural network models consistently outperformed persistent homology (PH) using SEDT filtrations with cubical complexes. While both methods exhibited increasing error as noise levels rose, the neural networks showed more gradual performance degradation. This suggests greater resilience to distortion and a stronger ability to infer structure from partially corrupted inputs.

This robustness stems from the networks' ability to learn contextual features. Unlike PH, which depends on hand-tuned filtration parameters and thresholds, neural networks learn data-driven mappings that integrate local and global cues, spatial geometry, and domain-specific patterns. This enables them to adapt to complex scenarios where the boundary between signal and noise is ambiguous.

A key advantage of the neural models is their ability to generalize across varying noise intensities within a single architecture. Trained on all noise levels simultaneously, they learn to infer the severity of degradation without external calibration or noise labels. This is particularly beneficial in real-world applications, where noise characteristics may be unknown or heterogeneous. For example, while PH requires manual parameter tuning to avoid mistaking Gaussian noise for structure, the network learns to internally disambiguate such cases.

Sequential fine-tuning further improved performance by allowing the network to recalibrate its learned representation to each dataset's topology, resolution, and geometry. However, this also introduced tradeoffs: previously learned representations were partially overwritten. For instance, while $\text{NN}_\text{V}$ performed well on Voronoi data, its performance declined after further training on DeePore and CEM500K (seen in $\text{NN}_\text{VDC}$). This highlights that notions of noise and structure are inherently context-dependent, and the model adapts its filters to the most recent domain.

Despite such shifts, the networks demonstrated useful generalization. The $\text{NN}_\text{VD}$ model, trained without access to CEM500K, still outperformed PH as noise increased. While its absolute MAE was higher than $\text{NN}_\text{VDC}$, the error increased smoothly across noise levels, suggesting that the discrepancy was due to domain unfamiliarity rather than noise sensitivity. This implies the model learned transferable structural priors, providing robustness even under domain shift. Notably, the $\text{NN}_\text{Simul}$ model, trained concurrently on all datasets, achieved comparable or better performance than PH at higher noise levels. Although it did not surpass the best fine-tuned models, its strong performance across datasets and robustness to degradation reinforce the viability of unified, noise-resilient estimators.

Ultimately, this study demonstrates that neural networks can serve as a reliable and scalable alternative to PH for estimating Betti numbers in noisy 2D binary images. Evaluated across synthetic and real-world datasets, our trained model consistently outperformed an optimally tuned PH baseline in terms of mean absolute error as noise increased. These results may inform future applications of TDA in domains where structure is partially degraded, and where traditional filtration-based methods may falter due to noise sensitivity or parameter dependence.

\begin{credits}
\subsubsection{\ackname} This research was supported by the Australian Government through the ARC's Discovery Projects funding scheme (project DP210103304). The first author was supported by a Research Training Program (RTP) Scholarship – Fee Offset by the Commonwealth Government.
\subsubsection{\discintname}
The authors declare no competing interests.
\end{credits}

\bibliographystyle{splncs04}
\bibliography{literature.bib}

\begin{thebibliography}{10}
\providecommand{\url}[1]{\texttt{#1}}
\providecommand{\urlprefix}{URL }
\providecommand{\doi}[1]{https://doi.org/#1}

\bibitem{Aurenhammer1991}
Aurenhammer, F.: Voronoi diagrams—a survey of a fundamental geometric data
  structure. ACM Computing Surveys  \textbf{23}(3),  345--405 (1991)

\bibitem{barnes2021comparative}
Barnes, D., Polanco, L., Perea, J.A.: A comparative study of machine learning
  methods for persistence diagrams. Frontiers in Artificial Intelligence
  \textbf{4},  681174 (2021)

\bibitem{ChazalMichel2021}
Chazal, F., Michel, B.: An introduction to topological data analysis:
  Fundamental and practical aspects for data scientists. Frontiers in
  Artificial Intelligence  \textbf{4} (2021). \doi{10.3389/frai.2021.667963}

\bibitem{chung2024morphological}
Chung, Y.M., Hu, C.S., Sun, E., Tseng, H.C.: Morphological multiparameter
  filtration and persistent homology in mitochondrial image analysis. Plos one
  \textbf{19}(9),  e0310157 (2024)

\bibitem{Cohen-SteinerEtAl2007}
Cohen-Steiner, D., Edelsbrunner, H., Harer, J.: Stability of persistence
  diagrams. Discrete \& Computational Geometry  \textbf{37}(1),  103--120
  (2007). \doi{10.1007/s00454-006-1276-5}

\bibitem{conrad2021cem500k}
Conrad, R., Narayan, K.: Cem500k, a large-scale heterogeneous unlabeled
  cellular electron microscopy image dataset for deep learning. Elife
  \textbf{10},  e65894 (2021)

\bibitem{EdelsbrunnerHarer2010}
Edelsbrunner, H., Harer, J.: Computational Topology: An Introduction. Applied
  Mathematics, American Mathematical Society (2010)

\bibitem{FasyEtAl2014}
Fasy, B.T., Lecci, F., Rinaldo, A., Wasserman, L., Balakrishnan, S., Singh, A.:
  Confidence sets for persistence diagrams. Annals of Statistics
  \textbf{42}(6),  2301--2339 (2014). \doi{10.1214/14-AOS1252}

\bibitem{HannouchChalup2023}
Hannouch, K.M., Chalup, S.: Learning to see topological properties in 4d using
  convolutional neural networks. In: Proceedings of 2nd Annual Workshop on
  Topology, Algebra, and Geometry in Machine Learning (TAG-ML), PMLR. vol.~221,
  pp. 437--454 (2023)

\bibitem{HannouchChalup2025}
Hannouch, K.M., Chalup, S.: Topology estimation of simulated 4d image data by
  combining downscaling and convolutional neural networks. ACM Transactions on
  Graphics  \textbf{44}(3) (2025). \doi{10.1145/3736717}

\bibitem{heiss2021impact}
Heiss, T., Tymochko, S., Story, B., Garin, A., Bui, H., Bleile, B., Robins, V.:
  The impact of changes in resolution on the persistent homology of images. In:
  2021 IEEE international conference on big data (big data). pp. 3824--3834.
  IEEE (2021)

\bibitem{HenselEtAl2021}
Hensel, F., Moor, M., Rieck, B.: A survey of topological machine learning
  methods. Frontiers in Artificial Intelligence  \textbf{4} (2021).
  \doi{10.3389/frai.2021.681108}

\bibitem{herring2019topological}
Herring, A., Robins, V., Sheppard, A.: Topological persistence for relating
  microstructure and capillary fluid trapping in sandstones. Water Resources
  Research  \textbf{55}(1),  555--573 (2019)

\bibitem{koseki2020assessment}
Koseki, K., Kawasaki, H., Atsugi, T., Nakanishi, M., Mizuno, M., Naru, E.,
  Ebihara, T., Amagai, M., Kawakami, E.: Assessment of skin barrier function
  using skin images with topological data analysis. NPJ systems biology and
  applications  \textbf{6}(1), ~40 (2020)

\bibitem{liu2022convnet}
Liu, Z., Mao, H., Wu, C.Y., Feichtenhofer, C., Darrell, T., Xie, S.: A convnet
  for the 2020s. Proceedings of the IEEE/CVF Conference on Computer Vision and
  Pattern Recognition (CVPR)  (2022)

\bibitem{Lloyd1982}
Lloyd, S.P.: Least squares quantization in pcm. IEEE Transactions on
  Information Theory  \textbf{28}(2),  129--137 (1982)

\bibitem{moon2020predicting}
Moon, C., Li, Q., Xiao, G.: Predicting survival outcomes using topological
  features of tumor pathology images. arXiv preprint arXiv:2012.12102  (2020)

\bibitem{Otsu1979}
Otsu, N.: A threshold selection method from gray-level histograms. IEEE
  Transactions on Systems, Man, and Cybernetics  \textbf{9}(1),  62--66 (1979)

\bibitem{OtterEtAl2017}
Otter, N., Porter, M.A., Tillmann, U., Grindrod, P., Harrington, H.A.: A
  roadmap for the computation of persistent homology. EPJ Data Science
  \textbf{6}(1), ~17 (2017)

\bibitem{paul2019estimating}
Paul, R., Chalup, S.: Estimating betti numbers using deep learning. In: 2019
  International Joint Conference on Neural Networks (IJCNN). pp.~1--7. IEEE
  (2019). \doi{10.1109/IJCNN.2019.8852013}

\bibitem{PeekEtAl2023}
Peek, D., Skerritt, M., Chalup, S.: Synthetic data generation and deep learning
  for the topological analysis of 3d data. In: International Conference on
  Digital Image Computing: Techniques and Applications (DICTA 2023). pp.
  121--128. IEEE (2023). \doi{10.1109/DICTA60407.2023.00025}, arXiv:2309.16968

\bibitem{Perlin1985}
Perlin, K.: An image synthesizer. In: Proceedings of the 12th Annual Conference
  on Computer Graphics and Interactive Techniques (SIGGRAPH). pp. 287--296
  (1985)

\bibitem{pritchard2023persistent}
Pritchard, Y., Sharma, A., Clarkin, C., Ogden, H., Mahajan, S.,
  S{\'a}nchez-Garc{\'\i}a, R.J.: Persistent homology analysis distinguishes
  pathological bone microstructure in non-linear microscopy images. Scientific
  Reports  \textbf{13}(1), ~2522 (2023)

\bibitem{rabbani2020deepore}
Rabbani, A., Babaei, M., Shams, R., Da~Wang, Y., Chung, T.: Deepore: A deep
  learning workflow for rapid and comprehensive characterization of porous
  materials. Advances in Water Resources  \textbf{146},  103787 (2020)

\bibitem{robins2016percolating}
Robins, V., Saadatfar, M., Delgado-Friedrichs, O., Sheppard, A.P.: Percolating
  length scales from topological persistence analysis of micro-ct images of
  porous materials. Water Resources Research  \textbf{52}(1),  315--329 (2016)

\bibitem{Serra1982}
Serra, J.: Image Analysis and Mathematical Morphology. Academic Press (1982)

\bibitem{singh2023topological}
Singh, Y., Farrelly, C.M., Hathaway, Q.A., Leiner, T., Jagtap, J., Carlsson,
  G.E., Erickson, B.J.: Topological data analysis in medical imaging: current
  state of the art. Insights into Imaging  \textbf{14}(1), ~58 (2023)

\bibitem{de2022ripsnet}
de~Surrel, T., Hensel, F., Carri{\`e}re, M., Lacombe, T., Ike, Y., Kurihara,
  H., Glisse, M., Chazal, F.: Ripsnet: a general architecture for fast and
  robust estimation of the persistent homology of point clouds. In:
  Topological, Algebraic and Geometric Learning Workshops 2022. pp. 96--106.
  PMLR (2022)

\bibitem{turkevs2021noise}
Turke{\v{s}}, R., Nys, J., Verdonck, T., Latr{\'e}, S.: Noise robustness of
  persistent homology on greyscale images, across filtrations and signatures.
  Plos one  \textbf{16}(9),  e0257215 (2021)

\bibitem{vipond2021multiparameter}
Vipond, O., Bull, J.A., Macklin, P.S., Tillmann, U., Pugh, C.W., Byrne, H.M.,
  Harrington, H.A.: Multiparameter persistent homology landscapes identify
  immune cell spatial patterns in tumors. Proceedings of the National Academy
  of Sciences  \textbf{118}(41),  e2102166118 (2021)

\end{thebibliography}

\end{document}